\documentclass[journal,twoside,web]{ieeecolor}
\usepackage{generic}
\usepackage{cite}
\usepackage{amsmath,amssymb,amsfonts}
\usepackage{algorithmic}
\usepackage{graphicx}
\usepackage{textcomp}
\usepackage[colorlinks, linkcolor=cyan, anchorcolor=cyan, citecolor=cyan]{hyperref}
\usepackage{multirow}
\usepackage{booktabs}

\usepackage{threeparttable}

\def\BibTeX{{\rm B\kern-.05em{\sc i\kern-.025em b}\kern-.08em
    T\kern-.1667em\lower.7ex\hbox{E}\kern-.125emX}}
\begin{document}
\title{Mind The Gap: Alleviating Local Imbalance for Unsupervised Cross-Modality Medical Image Segmentation} 

\author{ Zixian Su, Kai Yao,  Xi Yang, Qiufeng Wang, Yuyao Yan, Jie Sun and Kaizhu Huang
\thanks{ Zixian Su and Kai Yao are with both University of Liverpool and School of Advanced Technology, Xi'an Jiaotong-Liverpool University.}
\thanks{Xi Yang, Qiufeng Wang, Yuyao Yan and Jie Sun are with the School of Advanced Technology, Xi'an Jiaotong-Liverpool University, Suzhou, Jiangsu, 215000, China.}
\thanks{Kaizhu Huang is with Institute of Applied Physical Sciences and Engineering, Duke Kunshan University, Kunshan, Jiangsu, 215316, China. }
\thanks{Correspondence: xi.yang01@xjtlu.edu.cn and kaizhu.huang@dukekunshan.edu.cn} 
}
\maketitle

\begin{abstract}
Unsupervised cross-modality medical image adaptation aims to alleviate the severe domain gap between different imaging modalities without using the target domain label. 
A key in this campaign relies upon aligning the distributions of source and target domain. One common attempt is to enforce the global alignment between two domains, which, however, ignores the fatal local-imbalance domain gap problem, 
i.e., some local features with larger domain gap are harder to transfer. 
Recently, some methods conduct alignment focusing on local regions to improve the efficiency of model learning. While this operation may cause a deficiency of critical information from contexts.
To tackle this limitation, we propose a novel strategy to alleviate the domain gap imbalance considering the characteristics of medical images, namely \emph{Global-Local Union Alignment}. Specifically,
a feature-disentanglement style-transfer module first synthesizes the target-like source-content images to reduce the global domain gap. 
Then, a local feature mask is integrated to reduce the `inter-gap' for local features by prioritizing those discriminative features with larger domain gap. 
This combination of global and local alignment can precisely localize the crucial regions in segmentation target while preserving the overall semantic consistency.
We conduct a series of experiments with two cross-modality adaptation tasks, i,e. cardiac substructure and abdominal multi-organ segmentation. {Experimental results indicate that our method achieves state-of-the-art performance in both tasks. } 
\end{abstract}

\section{Introduction}
The success of deep learning models is highly dependent on the assumption that the training and testing data are i.i.d (independently identically distribution). 
However, in clinical practice, domain shift is widespread among different datasets due to the existence of multiple imaging modalities. Nevertheless, accurate segmentation can only be achieved with a large amount of labeled data in each modality, which is  tedious and time-consuming.
To tackle this problem, Unsupervised Domain Adaptation (UDA)~\cite{bengio2012deep,ganin2015unsupervised,zhu2017unpaired,ghifary2016deep} has been widely investigated to transfer the knowledge learned from a rich-labeled source domain to an unlabeled target domain, so as to train a model to segment the target images without labels.
\begin{figure}
  \centering
   \includegraphics[scale=0.33]{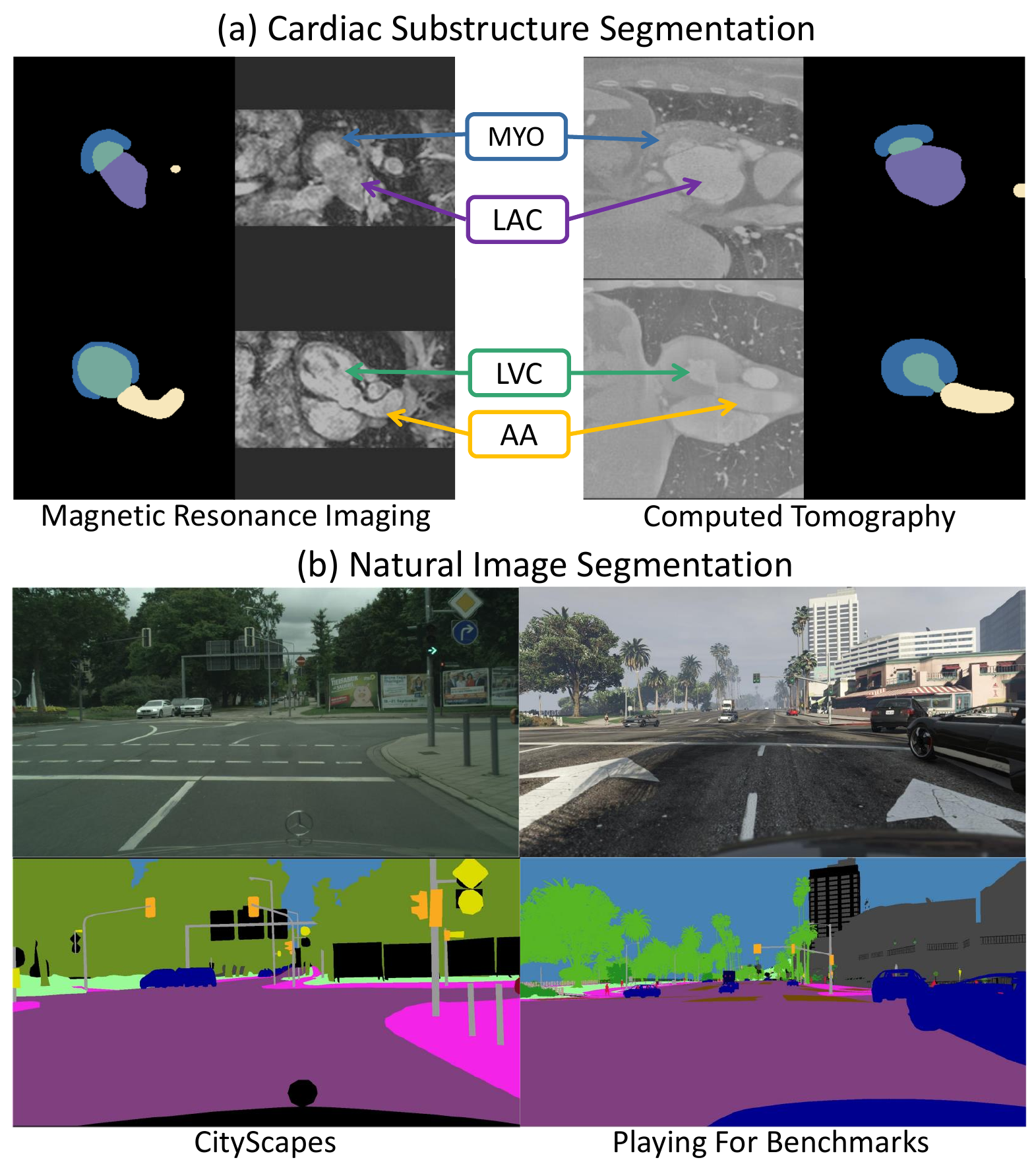}
  \caption{Severe domain gap can be seen in cross-modality cardiac dataset, as compared with natural images. Except for the style difference in two modalities, substurctures show diverse morphology. This phenomenon is more obvious for MYO and LVC, which indicates larger domain gap in these local regions.}
   \label{fig:HR}
\end{figure}


Although the existing UDA methods have shown impressive results in cross-modality medical image analysis~\cite{synseg,adaoutput,guo2021mcdcd}, most of them are under the assumption that all features share consistent domain gap. 
Though the key component of domain adaptation for semantic segmentation is quite straightforward - matching the overall distributions of two different domains, merely focusing on the global distribution gap while ignoring the `inter-gap' of local features appears inappropriate.

\begin{figure*}
  \centering
  
  \setlength{\belowcaptionskip}{-0.5cm}
   \includegraphics[scale=0.55]{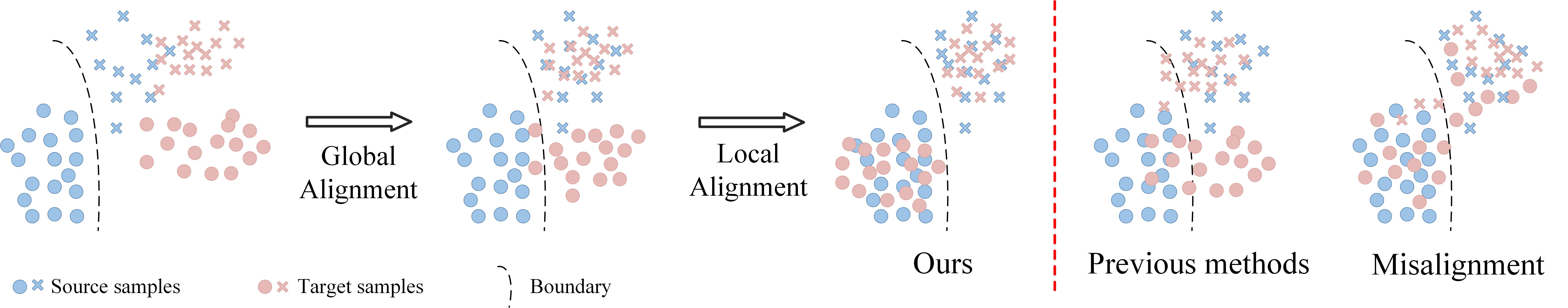}
      \caption{Illustration of our proposed method  vs.  global alignment}
   \label{fig:onecol1}
\end{figure*}
For an illustrative example shown in Fig.~\ref{fig:HR} (a), in cardiac segmentation of Magnetic Resonance Imaging (MRI) and Computed Tomography (CT) scans~\cite{CARDIACDATASET}, the transfer of knowledge is much easier for ascending aorta (AA) as its appearance is quite similar in both domains - they are both brighter than their surroundings, and the shape of this structure is close to each other in two domains. In contrast, the adaptation for the structure myocardium of the left ventricle (MYO) will be much harder. The segmentation task for this structure itself is challenging due to the fuzzy boundary, not to mention the difference in morphology and contrast between the two domains.
In this kind of scenario, if we only perform basic global alignment without considering the local gap imbalance, a possible consequence would be the over-alignment for the transferable features and insufficient adaptation for the discriminative features, as shown in Fig.~\ref{fig:onecol1}. Here, the transferable features indicate   those  with a relatively small domain gap (cross samples in Fig.~\ref{fig:onecol1}), while the discriminative features represent  the ones with a larger domain gap (circle samples in Fig.~\ref{fig:onecol1}). A bigger cost would be the misalignment of local features in order to achieve strict global alignment, in which the local features `inter-gaps' may even be enlarged. 

In~\cite{chen2020harmonizing}, Chen \emph{et al.} shared the similar view that strictly aligning the entire
feature distributions between domains is prone to result in negative transfer, since the transferability of different levels varies greatly. In addition, in~\cite{saito2019strong}, Saito \emph{et al.} identified an analogous phenomenon for object (instance-level) and scene layouts (image-level) in the task of object detection. Therefore, Chen \emph{et al.} proposed to handle a contradiction between transferability and discriminability globally and locally, while~\cite{saito2019strong} chose to leverage strong-local and weak-global alignment strategy instead.

In a similar spirit, we rethink the alignment strategy for domain gap problem in medical image data. In this study, our main goal is to improve the transferability of the segmentation object. Considering that the characteristics of medical images are quite different from those natural images~\cite{cordts2016cityscapes,Richter_2017} - the segmentation target is relatively small and centralized, while the background is comparably complex and different (see Fig.~\ref{fig:HR}~(a)(b)), the domain shift problem itself is more severe and challenging, as it implies that not only the segmentation target but also the irrelevant background has a large domain shift. Due to the one-of-a-kind existence of irrelevant and complex background, the performance of common attempts on natural datasets would be greatly reduced. In other words, the key issue is effective alignment in the segmentation target with local domain gap imbalance. Thus, we argue that global alignment as well as the local alignment are both essential for domain adaptation in medical images. Merely global alignment may hurt the overall performance,  potentially impairing the discriminability in target domain. While local alignment itself is insufficient to capture discriminative features in the segmentation target, it is likely to act on background area incorrectly under severe domain shift. Therefore, we first introduce a style-transfer module to pre-process the data beforehand to conquer the severe domain shift in the whole picture, as illustrated in Fig.~\ref{fig:onecol1}. This step proves vital, since it can make alignment at the domain level to globally alleviate the domain shift both in background and target, helping the downstream module prioritize the discriminative features with larger domain gap in the segmentation target instead of those in the whole picture. After that, the local feature masks are adopted in segmentation subnet produced by the uncertainty of the domain discriminator, which can be seen as the attention-like module that identifies the local domain gap in feature space unsupervisedly. This module could capture the discriminative features and 
emphasize these local alignment, ensuring the local-level semantic consistency for final prediction. 
To sum up, the framework is capable of paying more attention to local regions with larger variance in segmentation target instead of coarse global alignment. 
our main contributions are highlighted as follows:

1) We propose a novel strategy, Global-Local Union Alignment, which combines global alignment and local alignment while considering the characteristics of medical images, thereby promoting the overall transferability in cross-modality tasks with severe domain gap and local gap imbalance.

2) {The results in two bidirectional cross-modality segmentation tasks between MRI and CT: cardiac substructure segmentation~\cite{CARDIACDATASET} and abdominal multi-organ segmentation~\cite{emre2020chaos,ABDOMINALCT} demonstrate that our method achieves superior performance to state-of-the-art UDA methods generally.}

\section{Related Work}
\subsubsection{Domain Adaptation in Medical Image Analysis} 
There has been a rapid development of deep learning models for medical image analysis~\cite{med1,med2,med3,med4}. Nonetheless,  the effectiveness of deep learning model in the clinical application has been greatly hindered due to the requirement of a large amount of labelled data and the population of multi-modality datasets. To solve the domain shift problem, many domain adaptation methods have been proposed and achieved great performance in the field of medical image analysis~\cite{adaoutput,synseg,chen2020unsupervised}. For example, Wollmann \emph{et al.}~\cite{wollmann2018adversarial} propose a CycleGAN-based domain adaptation method for breast cancer classification. Similarly, this mode is also leveraged by Manakov \emph{et al.}~\cite{manakov2019noise} to tackle the retinal optical coherence tomography (OCT) image denoising problem. Mahmood \emph{et al.}~\cite{mahmood2018unsupervised} propose a reverse  unsupervised domain adaptation method to generate a synthetic-like representation of real endoscopy images for the notoriously difficult task of depth-estimation from endoscopy. Recently,  Chen \emph{et al.}~\cite{SIFA} present a novel UDA network for semantic segmentation which synergistically combines feature alignment and image alignment into one unified framework to improve the adaptation performance.  {Some of the latest work~\cite{liu2022margin,DSAN} even exploit networks pretrained on large-scale datasets which could leverage pretext knowledge to further boost performance on medical UDA tasks.}
\\

\subsubsection{Global Alignment vs. Local Alignment} The existing UDA methods can mainly be devided in three categories, adversarial training~\cite{adversarial1,adversarial2}, distance minimization~\cite{MDD,distance2} and self-training~\cite{self-training1,self-training2}. 
These classical methods align the global source and target domain distribution without considering the class/local-level distribution shift imbalance, which, however, leads some discriminable class-wise/local information to be confused.
Recently, researchers are increasingly investigating local-level alignment~\cite{luo2019taking,yu2021dast}, which performs local alignment of the source and target domains in the feature space leveraging the uncertainty of the model. These alignment methods have gained considerable performance improvement compared with the conventional ones in common datasets due to capturing the fine-grained domain information.
Besides, the combination of global and local alignment are undoubtedly arising research attention. Except for~\cite{saito2019strong,chen2020harmonizing} stated above which consider global and local alignment at multi-level feature space with various size of respective fields,~\cite{huang2020contextual} proposed contextual-relationships consistent domain adaptation via multi-scale entropy max-minimizing; and~\cite{qiang2021robust} adopted robust local preserving and global aligning network under the setting of noisy labels. 
Our framework is similar to~\cite{saito2019strong,chen2020harmonizing}  in that the local alignment module is introduced to reflect the discriminability and calibrate the transferability of feature representations.
 However, the proposed method is much different from  them both in global alignment strategy and the interaction method between global and local alignment.
~\cite{saito2019strong,chen2020harmonizing} make local alignment with a small respective field in feature space and global features (image-size) on deeper layers of the model, with both alignments unified in one stage. The global alignment in our work is operated on the whole distribution at domain level, which can be taken as a preprocessing step ahead of the local alignment. 

\section{Main Method}

\subsection{Method Overview}
\label{section:overview}

In this work, we focus on the UDA problem for medical image semantic segmentation, where the labeled source dataset $\left\{ x_s,y_s \right\}$ and unlabeled target set $\left\{ x_t \right\}$ are available. 
\begin{figure*}
  \centering
  \setlength{\belowcaptionskip}{-0.5cm}
   \includegraphics[scale=0.35]{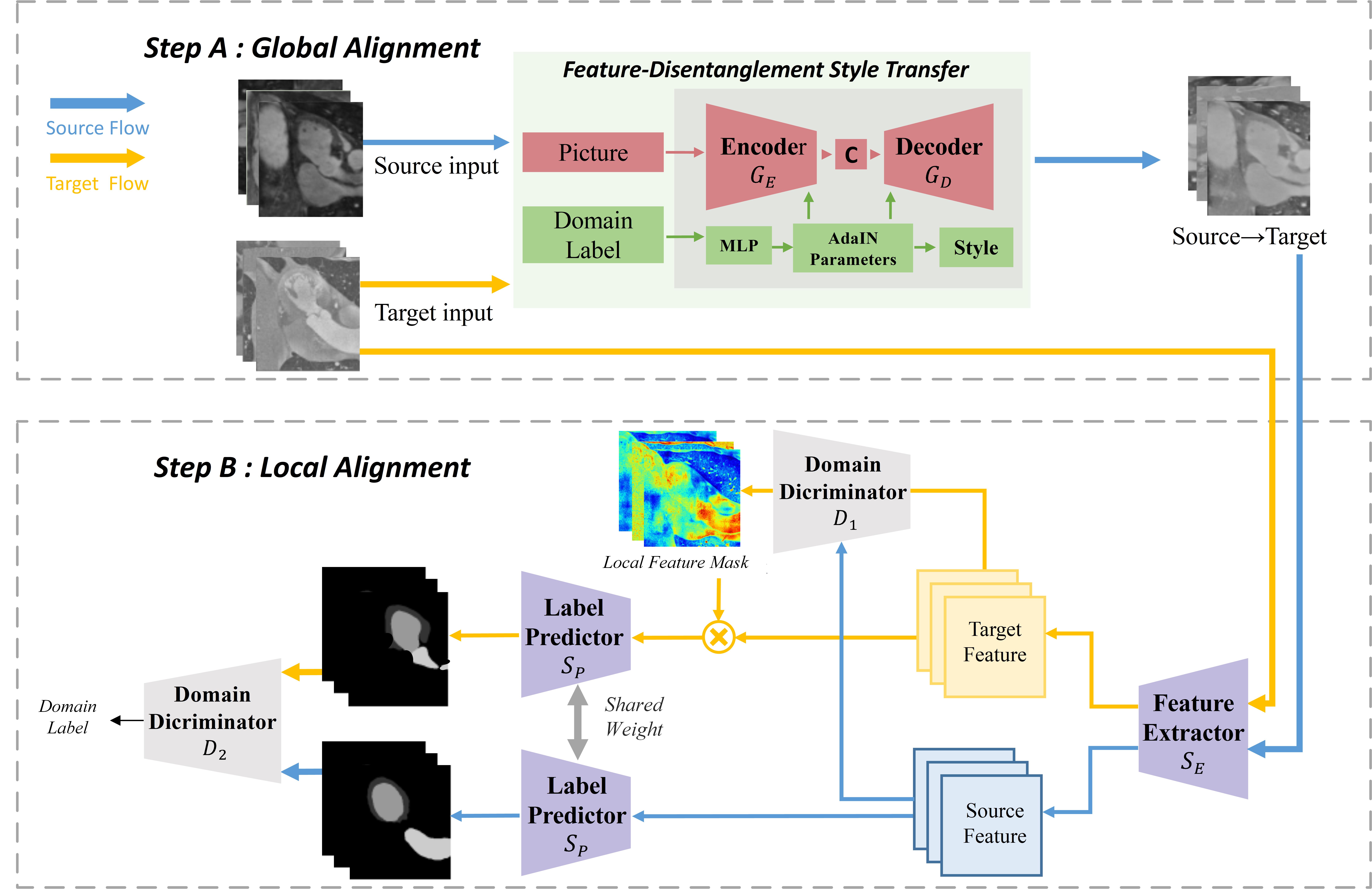}
  \caption{Overview of the whole framework. The feature-disentanglement style-transfer subnet first synthesizes the source-content target-style images. Combined with real target images, they are used to train the segmentation subnet, where a local feature mask is introduced to calibrate the discriminative areas.
The module $C$ in step A is the `content'(domain-invariant features) of an image. }
   \label{fig:onecol}
\end{figure*}

The overall framework is shown in Fig.~\ref{fig:onecol}. It is composed of a synthesis subnet and a segmentation subnet. Specifically, we first disentangle the features of two domains into a shared content space and individual style representations. By swapping the individual style representation to exchange the style of the original image, we decode them into style-transferred images while maintaining their anatomical structure. Then, the style-transferred source domain image $x_{s\rightarrow t}$ and the original target domain image $x_t$ are fed into the segmentation subnet. With a local feature mask computed by a pixel-wise domain discriminator, the target domain feature is reweighted and calibrated. 
The convolutional segmentor can attain more accurate prediction by concentrating on the discriminable features with larger domain shift, in order to balance the `inter-gap' between local features.\\ 

\subsubsection{Global Alignment: Feature-Disentanglement Synthesis Subnet} Similar to StarGAN~\cite{choi2018stargan}, we assume that the latent space of images can be decomposed into a content space and a style space. While in our framework, rather than taking a single domain as diverse style, we employ the unified domain style instead. Following this assumption, the image $x_i\in \mathcal{X}_i (i=s,t)$ can be represented as $x_i =(c_i,d_i)$, where $c_i$ is the content representation of image $x_i$ and $d_i$ is the style representation - domain label. Different from previous efforts that use an extra encoder to extract the domain-specific features to represent style, our work employs a multi-layer perceptron (MLP) to learn the style representation for each domain, which is represented by the domain label $d_i$. 
Inspired by some recent work that engages affine transformation parameters in normalization layers to represent styles, we equip both the decomposition and reconstruction processes with Adaptive Instance Normalization (AdaIN)~\cite{huang2017arbitrary} layers to introduce the style representations.

 As shown in Fig.~\ref{fig:onecol}, we exploit a 2D auto-encoder architecture as the synthesis subnet $\mathcal{G}$, which consists of a shared encoder $\mathcal{G}_{E}$ and decoder $\mathcal{G}_{D}$. First, the content representation $c_x$ of the image $x$ is obtained using decoder $\mathcal{G}_{E}$ conditioned on the domain label $d_i$, i.e., $c_i = \mathcal{G}_E(x_i,d_i)$. Then, we compose the reconstructed image $x_{ii}$ based on the content $c_x$ and domain label $d_i$, i.e., $x_{ii} = \mathcal{G}_D(\mathcal{G}_E((x_i,d_i)),d_i)$. Similarly, the style-transferred image (from domain $i$ to domain $j$) can be integrated as $x_{ij} = \mathcal{G}_D(\mathcal{G}_E((x_i,d_i)),d_j), i\neq j$. In addition, a Markovian discriminator~\cite{isola2017image} $D$ is introduced as the image discriminator for both domains, which learns to determine whether an image $x_i$ is a real image of its domain $\mathcal{X}_i$ or a fake one generated by $\mathcal{G}$, i.e., $D_i(x_i) = D(x_i|d_i)$. \\
 
\subsubsection{Local Alignment: Segmentation Subnet with Dual Discriminators}
As shown in Fig.~\ref{fig:onecol}, local alignment stage is mainly composed of a segmentation network $\mathcal{S}$ and two discriminator networks $D_{1},D_2$. For better description, the segmentation network can be further decomposed into two compositions, a feature extractor $\mathcal{S}_E$ and a label predictor $\mathcal{S}_P$. We adopt Deeplab-ResNet50~\cite{chen2017deeplab} as the segmentation backbone. Discriminators are two convolutional classifiers working in feature space and image space respectively.
The target-like source-content images $x_{s\rightarrow t}$ will be first extracted by $\mathcal{S}_E$ to form the feature representation $f_{s\rightarrow t}$, i.e., $f_{s\rightarrow t} = \mathcal{S}_E(x_{s\rightarrow t})$. Then the feature map $f_{s\rightarrow t}$ will be taken into the label predictor $S_P$ for semantic segmentation to get the final result $p_{s\rightarrow t}$. Meanwhile, the feature map $f_{s\rightarrow t}$ and final prediction $p_{s\rightarrow t}$ are input into the discriminators as the source samples of adversarial training, respectively.

For the real target flow, the feature map $f_t$ extracted by $\mathcal{S}_E$ is first fed into $D_1$. $D_1$ will evaluate the feature distribution of target data $f_t$ and produce a local feature mask $m$, where $m = |D_{1}(f_t)|$. The feature mask then reweights the original feature map $f_t$ to form a new feature map $\hat{f_t} = f_t + f_t\otimes expand(tanh(m))$ and feeds it into $\mathcal{S}_P$ to yield the final prediction. Here, $expand$ operation replicates the local feature map with channel $1$ to a channel dimension that is the same as the input feature for the subsequent element-wise multiplication and $tanh$ is introduced as a normalization layer to prevent the gradient exploration in the early training stage. For the second discriminator $D_2$, it evaluates the final predictions calibrated through the feature mask $m$, and further narrows the pixel-level domain gap between source prediction $p_s$ and target prediction $p_t$ through adversarial loss.

\subsection{Objective Function}\label{section:losses}
The proposed model is featured in two stages. In stage one, the training loss consists of reconstruction loss, cycle consistency loss, and adversarial loss.
First of all, the reconstruction loss $\mathcal{L}_{rec}$ is a promise for the generator $\mathcal{G}$ to rebuild the image $x_{i}$ and content $c_i$ during cross-domain translation:
\begin{equation*}
    \mathcal{L}_{rec\_im}(\mathcal{G}_{E},\mathcal{G}_{D}) = \mathbb{E}_{x_i\sim p(X_i)}[\|\mathcal{G}_{D}(\mathcal{G}_{E}(x_i,d_i),d_i)-x_i\|_1]\;,
\end{equation*}
\begin{equation}
\nonumber
    \mathcal{L}_{rec\_c}(\mathcal{G}_{E},\mathcal{G}_{D}) = \mathbb{E}_{c_i}[\|\mathcal{G}_{E}(\mathcal{G}_{D}(c_i,d_j),d_j)-c_i\|_1]\;.
\end{equation}
Meanwhile, a cycle-consistency loss is adopted to guarantee that the mapping between the original image and the generated image is a unique cross-domain. Namely, for a given image $x_i$, it should return to itself when translated $\mathcal{X}_i \rightarrow \mathcal{X}_j \rightarrow \mathcal{X}_i$. The cycle-consistency loss is described as:
\begin{equation*}
\begin{split}
    &\mathcal{L}_{cyc}(\mathcal{G}_{E},\mathcal{G}_{D})=
    \\&\mathbb{E}_{x_i\sim p(X_i)}[\|\mathcal{G}_{D}(\mathcal{G}_{E}(\mathcal{G}_{D}((\mathcal{G}_{E}(x_i,d_i),d_i),d_j),d_i)-x_i\|_1]\;.
\end{split}
\end{equation*}

\noindent Moreover, we would like to make sure that the data $x_{ij}$ generated by $\mathcal{G}$ is as close to the real data distribution as possible, an adversarial loss is included in GAN training:
\begin{equation}
\nonumber
\begin{split}
    \mathcal{L}_{adv}(\mathcal{G},D) =&
     \mathbb{E}_{x_j\sim p(X_j)}[\log D((x_j|d_j)]+ \\&
    \mathbb{E}_{x_i\sim p(X_i)}[1 - \log D(\mathcal{G}_{D}(\mathcal{G}_{E}(x_i,d_i),d_j)|d_j)]\;.
\end{split}
\end{equation}
Finally, we jointly train the encoder, decoder, and domain discriminator, which is shown as:
\begin{equation}
\begin{split}
\nonumber
\mathop{\min}_{\mathcal{G}_{E},\mathcal{G}_{D}}\mathop{\max}_{\mathcal{D}}\ &\lambda_{rec\_im}\mathcal{L}_{rec\_im} + \lambda_{rec\_c}\mathcal{L}_{rec\_c}
 \\& + \lambda_{cyc}\mathcal{L}_{cyc} +\lambda_{adv}\mathcal{L}_{adv}.
\end{split}
\end{equation}

\noindent In stage two, the training loss is composed of the segmnetation loss and adversarial loss.
Segmentation loss $\mathcal{L}_{seg}$ is for segmentor $S(\mathcal{S}_E,\mathcal{S}_P)$ to output more accurate prediction  in a supervised way:
\begin{equation}
\nonumber
\begin{split}
    \mathcal{L}_{seg}(\mathcal{S}_E,\mathcal{S}_P) =&  \mathbb{E}_{x_i,y_i\sim p(X_S,Y_S)}(\ell_{ce}(\mathcal{S}(x_i),y_i) \\&+ \ell_{dice}(\mathcal{S}(x_i),y_i))\;.
\end{split}
\end{equation}
$\ell_{ce}$ and $\ell_{dice}$ term represent the multi-class cross-entropy loss and the Dice loss. 

The adversarial loss can be decomposed into two parts,  $\mathcal{L}_{adv}^{D_1}(\mathcal{S}_E,D_1)$ and $\mathcal{L}_{adv}^{D_2}(\mathcal{S}_E,\mathcal{S}_P,D_2)$, indicating the adversarial learning losses of the two discriminators, respectively. The process is summarized as below:
\begin{equation}
\begin{split}
\nonumber
   \mathcal{L}_{adv}^{D_1}(\mathcal{S}_E,D_1) =& \mathbb{E}_{x_{s\rightarrow t}\sim p(X_{S\rightarrow T})}[\log D_{1}(\mathcal{S}_E(x_{s\rightarrow t})] \\&+ \mathbb{E}_{x_t\sim p(X_T)}[1 - \log D_{1}(\mathcal{S}_E(x_t))]\;.
\end{split}
\end{equation}
\begin{equation}
\begin{split}
\nonumber
   \mathcal{L}_{adv}^{D_2}(\mathcal{S},D_2) =& \mathbb{E}_{x_{s\rightarrow t}\sim p(X_{S\rightarrow T})}[\log D_{2}(\mathcal{S}(x_{s\rightarrow t})] \\&+ \mathbb{E}_{x_t\sim p(X_T)}[1 - \log D_{2}(\mathcal{S}(x_t))]\;.
\end{split}
\end{equation}

The overall training objective function of the segmentation subnet is:
\begin{equation}
\begin{split}
\nonumber
\min\limits_{\mathcal{S}_E,\mathcal{S}_P}\max\limits_{D_1,D_2}\  &\mathcal{L}_{seg}(\mathcal{S}_E,\mathcal{S}_P) + \lambda_{d_1}\mathcal{L}_{adv}^{D_1}(\mathcal{S}_E,D_1) \\&+ \lambda_{d_2}\mathcal{L}_{adv}^{D_2}(\mathcal{S}_E,\mathcal{S}_P,D_2)\;.
\end{split}
\end{equation}
$\lambda_{rec\_im},$$\lambda_{cyc},$$\lambda_{rec\_c},$$\lambda_{adv},$$\lambda_{d_1},$$\lambda_{d_2}$ are introduced to balance the relative importance of different terms.

\section{Experiments}
\subsection{Setup}
\subsubsection{Implementation Details} Our model is implemented using PyTorch toolbox on a Quadro RTX 8000 (48GB memory). In style-transfer stage, ADAM solver~\cite{kingma2014adam} is adopted with a batch size of 8. Learning rate is set to $0.0002$, and weight decay is $0.0001$. In the segmentation stage,  we choose SGD algorithm as the optimizer for segmentator $\mathcal{S}$ with momentum of $ 0.9$ and weight decay of $0.0005$, while for the dual discriminators we use ADAM optimizer with $\beta_1 = 0.9$ and $\beta_2 = 0.99$. The weight decay method is polynomial decay with a power of $0.9$. All the networks are trained from scratch. We tune the batch size and learning rate empirically in each task. For the hyper-parameters,  $\lambda_{rec\_im}=\lambda_{cyc}=20$, $\lambda_{rec\_c}=\lambda_{adv}=1$, $\lambda_{d_1}=\lambda_{d_2}=0.01$. We train the style-transfer model for 50k iterations.  After that, the pre-processed training set is fed into the segmentation network. The validation set is used to find the model with the  best performance. \\

\subsubsection{Datasets} Two challenging adaptation tasks, e.g., Cardiac substructure segmentation and Abdominal multi-organ segmentation, are performed in the experiment. 1) Cardiac substructure segmentation utilizes the Multi-Modality Whole heart Segmentation (MMWHS) Challenge 2017 dataset~\cite{CARDIACDATASET}, which is composed of 20 labelled MRI volumes and 20 labelled CT volumes. The goal is to parse the four heart structures: the ascending aorta (AA), the left atrium cavity (LAC), the left ventricle blood cavity (LVC) and the myocardium of the left ventricle (MYO).  2) Abdominal multi-organ segmentation is implemented with 30 volumes CT data from~\cite{ABDOMINALCT} and 20 volumes of T2-SPIR MRI data from ISBI CHAOS Challenge 2019~\cite{emre2020chaos} with both groundtruth masks provided. We aim to segment the four key organs, including liver, right kidney, left kidney and spleen.

The pre-processing includes normalization, random crop, rotation and other augmentation operations. The size for each slice is $256\times 256$ in the cardiac dataset and $512 \times 512$ in the abdominal dataset. For cardiac dataset, we used the pre-processed version from~\cite{SIFA}.  It is worth noting that two domains of pictures in both tasks are unpaired. \\

\subsubsection{Evaluation Indicators} Two metrics commonly used in medical image segmentation are adopted: Dice similarity coefficient (Dice) and the average surface distance (ASD). Dice score measures the voxel-level intersection part between the prediction mask and the ground truth mask, while the ASD coefficient represents the average distances between the surfaces of the two in 3D.

\subsection{Experimental Results}

We  evaluate  quantitatively and qualitatively the proposed method against other SOTAs including  CycleGAN~\cite{zhu2017unpaired}, CyCADA~\cite{hoffman2018cycada}, SynSeg-Net~\cite{synseg}, SIFA~\cite{SIFA}, DSFN~\cite{DSFN}, UESM~\cite{bian2020uncertainty}, DSAN~\cite{DSAN}, CUDA~\cite{du2021constraint}, DDA-GAN~\cite{DDAGAN} and UMDA~\cite{UMDA}. Among them, the earlier methods CycleGAN and SynSeg-Net conduct adversarial alignment at the pixel level, while the remaining methods exploit both image and feature alignment.  

\begin{figure*}
  \centering
   \includegraphics[scale=0.55]{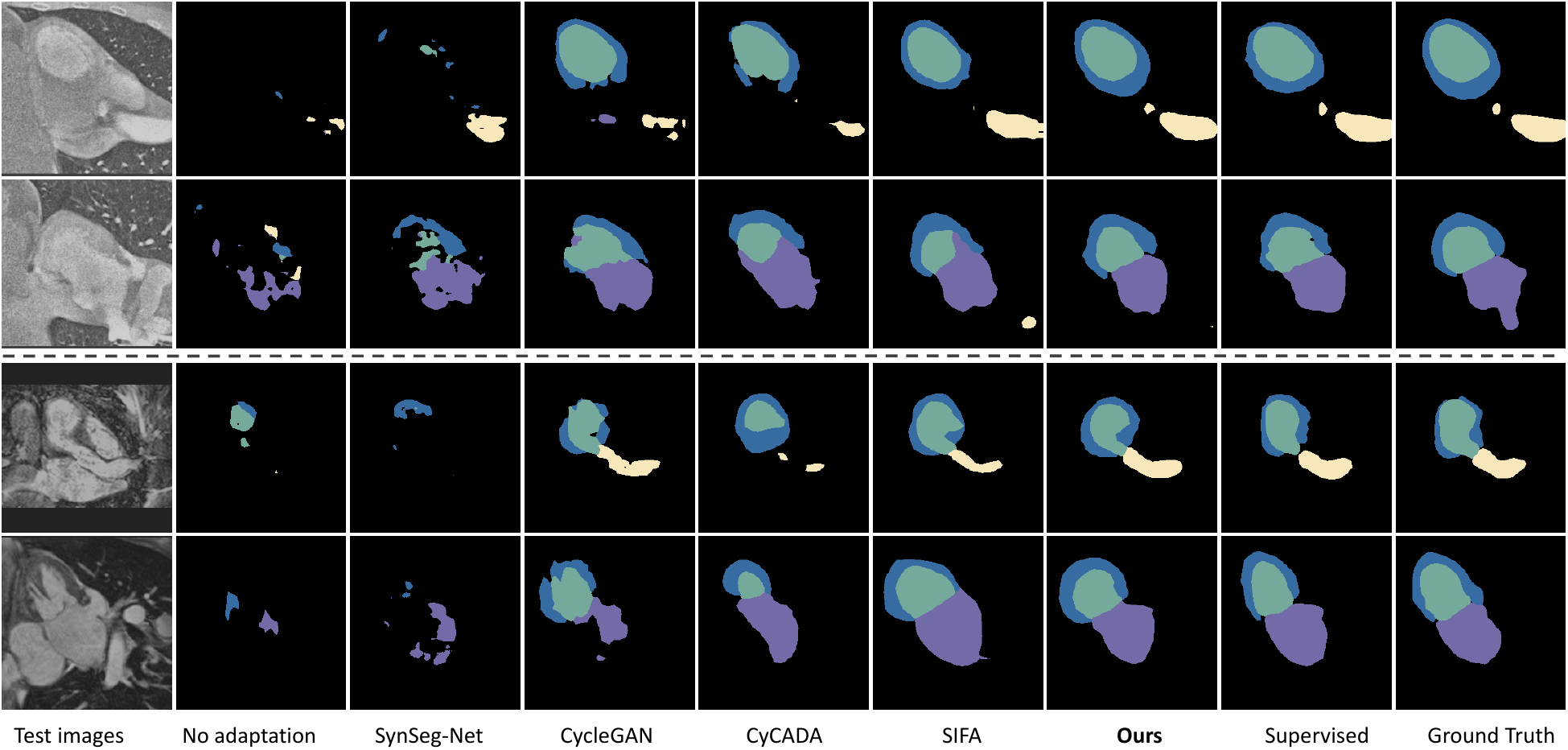}
   \caption{Qualitative comparison result of cardiac segmentation. The first two rows are CT images and the following two are MR images. The cardiac substructure, AA, LAC, LVC, MYO are marked in yellow, purple, green and blue, respectively. }
   \label{fig:cardiacvis}
\end{figure*}

\begin{figure*}

  \centering
  \includegraphics[scale=0.55]{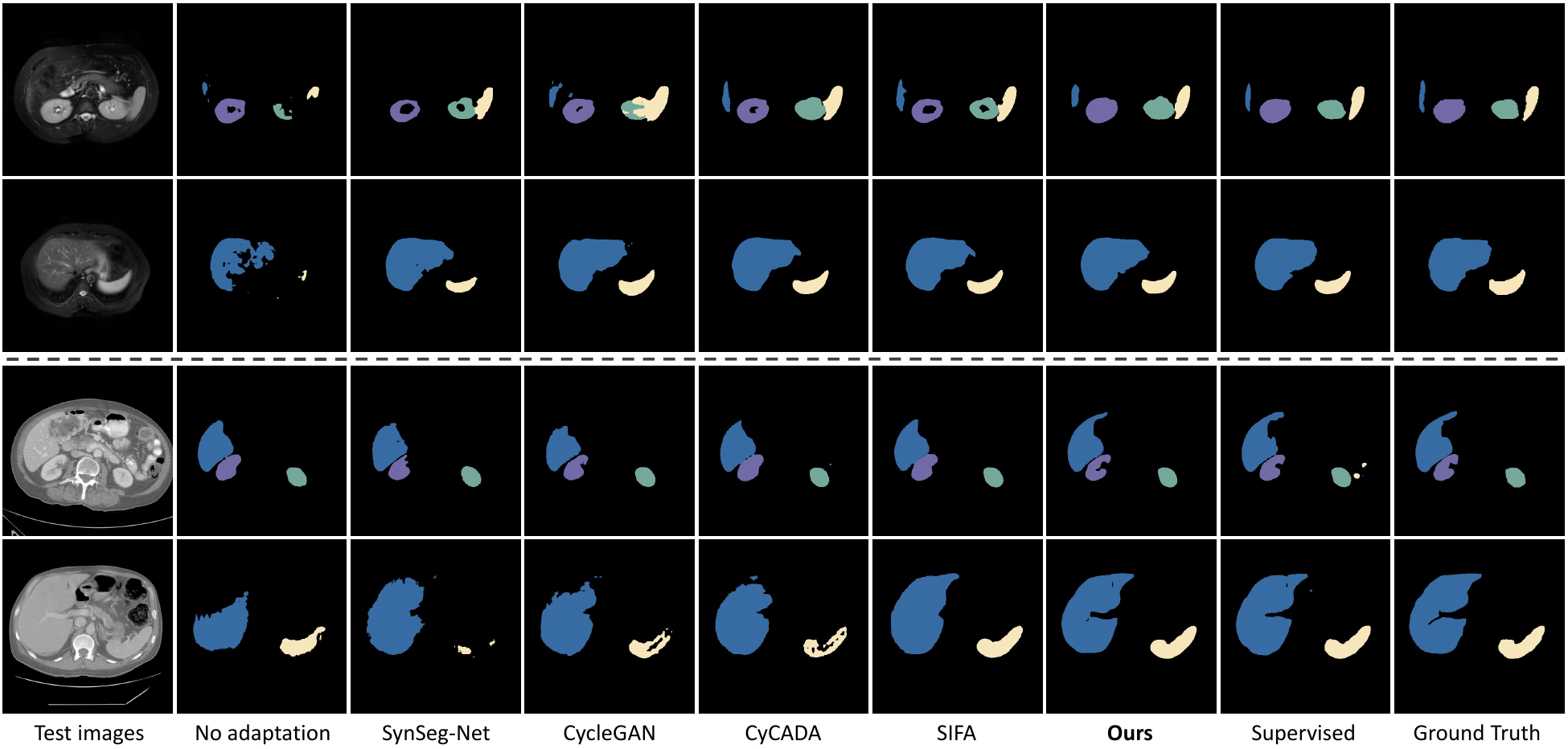}
  \caption{Qualitative comparison results of abdominal segmentation. The first two rows are MR images, and the following two are CT images. The organs, Liver, Right kidney, Left kidney and Spleen, are marked in blue, purple, green and yellow, respectively. }
  \label{fig:abdominalvis}
\end{figure*}

Table~\ref{cardiactable}, \ref{abdominaltable} and Fig.~\ref{fig:cardiacvis}, \ref{fig:abdominalvis} respectively show the comparison results in cardiac segmentation and abdominal multi-organ segmentation.
{Part of the lines are absent due to the lack of public codes.
We conduct CycleGAN, SynSeg-Net and CyCADA on our datasets to obtain the final numbers. Performance of the remaining methods come from their papers. It is also noted that all the reported results use the unpretrained backbones for fair comparison.}
 We also provide the results without adaptation and supervised training for comparison, where only a segmentation backbone is adopted. These two results can be approximately regarded as the lower and upper bounds for UDA. 

From these results, it can be seen that 1) There is a drastic performance drop between no adaptation and supervised training, indicating that a large domain gap exists between two domains in both tasks. 
2) {Our proposed method outperforms current state-of-the-art methods generally. }Compared with CyCADA~\cite{hoffman2018cycada}, which takes a similar two-stage setting but uses conducts global alignment only, our method improves the Dice score from 56.55\% to 82.42\% in Cardiac  MRI$\rightarrow$CT segmentation. 
{Although our method (82.42\%) is slightly inferior to CUDA (82.90\%) in task Caridac MRI-CT, it exceeds CUDA by a large margin in the inversed direction (Ours 69.78\% vs. CUDA 65.50\%). The main reason may be the superiority of their segmentation backbone, which accounts for the main contributions of their work. Moreover, the results on  abdominal dataset further corroborate the effectiveness of our method.}
3) Our proposed method appears to perform better in classes with larger domain gaps. Regions with lower transferability are more likely to appear in classes with larger domain gap, which was empirically derived from our observations of the local feature mask results. As in MYO of Cardiac MRI$\rightarrow$ CT and LVC of CT$\rightarrow$MR, our method exhibits more improvements than other classes in the same task.

\begin{table*}[]
\centering
\scriptsize
\renewcommand{\arraystretch}{1.3}
\setlength\tabcolsep{2.5pt}
\caption{Performance Comparison of Different Methods for  Cross-modality Cardiac Segmentation}
\centering
\begin{tabular}{ccccccccccc||cccccccccclllll}
\cline{1-22}
                                             & \multicolumn{10}{c||}{\textbf{Cardiac MRI - CT}}                                                        & \multicolumn{10}{c}{\textbf{Cardiac CT - MRI}}                                                         &                      &                      &                      &                      \\ 
\cline{1-22}

\multicolumn{1}{c|}{\multirow{2}{*}{Method}} & \multicolumn{5}{c|}{Dice(\%) $\uparrow$}                                & \multicolumn{5}{c||}{ASD(voxel) $\downarrow$}         & \multicolumn{5}{c|}{Dice(\%) $\uparrow$}                                & \multicolumn{5}{c}{ASD(voxel) $\downarrow$}          & \multicolumn{1}{c}{} & \multicolumn{1}{c}{} & \multicolumn{1}{c}{} & \multicolumn{1}{c}{} \\ 
\cline{2-22}
\multicolumn{1}{c|}{}            &AA    & LAC   & LVC   & MYO   & \multicolumn{1}{c|}{Avg} & AA    & LAC   & LVC   & MYO   & Avg & AA    & LAC   & LVC   & MYO   & \multicolumn{1}{c|}{Avg} & AA    & LAC   & LVC   & MYO   & Avg &                      &                      &                      &                      \\
\cline{1-22}
\multicolumn{1}{c|}{Supervised training}    & 88.49 & 87.71 & 90.69 & 83.33 & \multicolumn{1}{c|}{87.55}   & 3.64  & 3.19  & 2.03  & 2.14  & 2.75    & 83.61 & 84.16 & 90.74 & 76.40 & \multicolumn{1}{c|}{83.73}   & 1.98  & 1.73  & 1.58  & 1.64  & 1.73    &                      &                      &                      &                      \\
\multicolumn{1}{c|}{No adaptation}         & 30.75 & 26.25 & 10.10 & 1.36  & \multicolumn{1}{c|}{17.12}   & 26.92 & 18.45 & 16.75 & 28.78 & 22.73   & 12.79 & 11.52 & 10.21 & 8.61  & \multicolumn{1}{c|}{10.78}   & 11.74 & 11.21 & 14.10 & 11.23 & 12.07   &                      &                      &                      &                      \\ \cline{1-1}\cline{2-22}
\multicolumn{1}{c|}{CycleGAN\cite{zhu2017unpaired}}               & 72.58 & 71.88 & 50.45 & 31.52 & \multicolumn{1}{c|}{56.61}   & 12.50 & 14.76 & 9.08  & 8.48  & 11.21   & 61.90 & 40.97 & 60.27 & 41.59 & \multicolumn{1}{c|}{51.18}   & 5.83  & 8.43  & 7.84  & 5.12  & 6.81    &                      &                      &                      &                      \\
 \multicolumn{1}{c|}{SynSeg-Net\cite{synseg}}     & 66.99 & 62.11 & 47.20 & 38.00 & \multicolumn{1}{c|}{53.58}   & 12.85 & 8.19  & 7.31  & 10.22  & 9.64     & 58.50 & 39.09 & 62.49 & 31.10 & \multicolumn{1}{c|}{47.80}   & 9.55  & 10.03 & 7.81  & 5.04  & 8.10    &                      &                      &                      &                      \\
  \multicolumn{1}{c|}{CyCADA\cite{hoffman2018cycada}}       & 68.18 & 64.23 & 48.90 & 44.89 & \multicolumn{1}{c|}{56.55}   & 9.69  & 10.36  & 9.41  & 11.77 & 10.34    & 59.65 & 42.77 & 62.21 & 36.90 & \multicolumn{1}{c|}{50.38}   & 8.40  & 12.19 & 6.09  & 6.26  & 8.23  &                      &                      &                      &                      \\
\multicolumn{1}{c|}{SIFA\cite{chen2020unsupervised}    }          & 81.30 & 79.50 & 73.80 & 61.60 & \multicolumn{1}{c|}{74.10}   & 7.90  & 6.20  & 5.50  & 8.50  & 7.00    & 65.30 & 62.30 & 78.90 & 47.30 & \multicolumn{1}{c|}{63.40}   & 7.30  & 7.40  & 3.80  & 4.40  & 5.70    &                      &                      &                      &                      \\
\multicolumn{1}{c|}{DSFN\cite{DSFN}}               & 84.70 & 76.90 & 79.10 & 62.40 & \multicolumn{1}{c|}{75.80}   & -   & -   & -   & -   & -     & -     & -     & -     & -     & \multicolumn{1}{c|}{-}       & -     & -     & -     & -     & -       &                      &                      &                      &                     
\\


 \multicolumn{1}{c|}{DDA-GAN\cite{DDAGAN}}            & 68.30 & 75.70 & 78.50 & 77.80 & \multicolumn{1}{c|}{75.10}   & 6.50  & 4.80  & 5.40  & 5.20  & 5.50    & - & - &- &-&\multicolumn{1}{c|}{-}  & - & - & -  & - & -  &                      &                      &                      &       \\ 
 \multicolumn{1}{c|}{UMDA\cite{UMDA}}        & \textbf{89.20} & 82.70 &  82.60 & 66.20 & \multicolumn{1}{c|}{80.20}   & 6.70  & 3.60  & 4.50  & 3.00  & 4.00    & - & - &- &-&\multicolumn{1}{c|}{-}  & - & - & -  & - & -  &                      &                      &                      &       \\ 
 \multicolumn{1}{c|}{UESM\cite{bian2020uncertainty}}                   & 84.15 & 88.30& 84.32 & 71.42 & \multicolumn{1}{c|}{82.05}   & \textbf{3.87}  & 3.49  & 3.81  & 3.70  & \textbf{3.71}    & - & - & - & - & \multicolumn{1}{c|}{-}   & \textbf{-}  & -  & -  & -  & -    &                      &                      &                      &                      \\
 \multicolumn{1}{c|}{DSAN\cite{DSAN}}               & 79.92 & 84.76 & 82.77 & 66.52 & \multicolumn{1}{c|}{78.50}   & 7.68  & 6.65  & 3.77  & 5.59  & 5.92    & \textbf{71.29} & 66.23 & 76.20 & 52.07 & \multicolumn{1}{c|}{66.45}   & \textbf{4.44}  & 7.30  & 5.46  & 4.25  & 5.36    &                      &                      &                      &                      \\

\multicolumn{1}{c|}{CUDA\cite{du2021constraint}}       & {87.20} & \textbf{88.50} & 83.00 & \textbf{72.80} & \multicolumn{1}{c|}{\textbf{82.90} }  & 7.03  & \textbf{2.80}  & 5.20  & 6.80  & 5.50    & 68.50 & 66.10 &79.10 &42.80&\multicolumn{1}{c|}{65.50}  & 5.10 & 4.10 & 3.20  & 4.10  & 4.10  &                      &                      &                      &       \\ 
\cline{1-22}
\multicolumn{1}{c|}{\textbf{Ours}}               & 86.43 & {86.35} & \textbf{84.97} & 71.92 & \multicolumn{1}{c|}{82.42}   & 4.78  & 3.29  & \textbf{3.68}  & \textbf{3.35}  & 3.77    & 70.20 & \textbf{70.54} & \textbf{85.27} & \textbf{53.10} & \multicolumn{1}{c|}{\textbf{69.78}}   & 5.76  & \textbf{3.63}  & \textbf{2.84}  & \textbf{3.49}  & \textbf{3.93}    &                      &                      &                      &                      \\ 
\cline{1-22}
\end{tabular}

\label{cardiactable}
\end{table*}

\begin{table*}[]
\centering
\scriptsize
\renewcommand{\arraystretch}{1.3}
\setlength\tabcolsep{2.7pt}
\centering
\caption{Performance Comparison of Different Methods for  Cross-modality Abdominal Segmentation}
\begin{tabular}{ccccccccccc||ccccccccccllll}
\cline{1-21}
                                             & \multicolumn{10}{c||}{\textbf{Abdominal MRI - CT}}                                                                    & \multicolumn{10}{c}{\textbf{Abdominal CT - MRI}}                                                                     &                      &                      &                      &                      \\ \cline{1-21}
\multicolumn{1}{c|}{\multirow{2}{*}{Method}} & \multicolumn{5}{c|}{Dice(\%) $\uparrow$}                                       & \multicolumn{5}{c||}{ASD(voxel) $\downarrow$}                & \multicolumn{5}{c|}{Dice(\%) $\uparrow$}                                       & \multicolumn{5}{c}{ASD(voxel) $\downarrow$}                 & \multicolumn{1}{c}{} & \multicolumn{1}{c}{} & \multicolumn{1}{c}{} & \multicolumn{1}{c}{} \\ \cline{2-21}
\multicolumn{1}{c|}{}                        & Liver & R.kid & L.Kid & Spleen & \multicolumn{1}{c|}{Avg} & Liver & R.kid & L.Kid & Spleen & Avg & Liver & R.kid & L.Kid & Spleen & \multicolumn{1}{c|}{Avg} & Liver & R.kid & L.Kid & Spleen & Avg &                      &                      &                      &                      \\ \cline{1-21}
\multicolumn{1}{c|}{Supervised training}     & 95.08 & 91.06    & 93.62    & 89.37  & \multicolumn{1}{c|}{92.28}   & 0.66  & 1.18     & 0.73     & 1.11   & 0.92   & 93.21 & 93.42    & 93.22    & 89.59  & \multicolumn{1}{c|}{92.36}   & 0.70  & 0.48     & 0.55     & 1.12   & 0.71    &                      &                      &                      &                      \\
\multicolumn{1}{c|}{No adaptation}           & 78.65 & 54.52    & 47.84    & 55.48  & \multicolumn{1}{c|}{59.12}   & 3.86 & 8.72 & 3.19 & 4.88 & 5.16    & 58.30 & 61.08    & 59.85    & 73.95  & \multicolumn{1}{c|}{63.30}   & 3.93  & 1.82     & 2.58     & 2.42   & 2.69    &                      &                      &                      &                      \\ \cline{1-21}
\multicolumn{1}{c|}{SynSeg-Net\cite{synseg}}              & 82.15 & 79.36    & 68.06    & 80.44  & \multicolumn{1}{c|}{77.50}   & 2.67  & 1.98     & 2.09     & 2.65   & 2.34    & 84.30 & 88.61    & 71.92    & 74.39  & \multicolumn{1}{c|}{79.81}   & 2.94  & 1.86     & 2.33     & 2.51   & 2.41    &                      &                      &                      &                      \\
\multicolumn{1}{c|}{CyCADA\cite{hoffman2018cycada}}                  & 81.08 & 75.95   & 73.69    & 76.13  & \multicolumn{1}{c|}{76.71}   & 2.24  & 1.96     & 1.94     & 1.74   & 1.97    & 85.72 & 88.56    & 75.51    & 78.46  & \multicolumn{1}{c|}{82.06}   & 2.74  & 2.16     & 1.91     & 2.29   & 2.28    &                      &                      &                      &                      \\
\multicolumn{1}{c|}{SIFA\cite{chen2020unsupervised}}                   & 88.00 & 83.30    & 80.90    & 82.60  & \multicolumn{1}{c|}{83.70}   & \textbf{1.20}  & \textbf{1.00}     & \textbf{1.50}     & \textbf{1.60}   & \textbf{1.30}    & 90.00 & 89.10    & 80.20    & 82.30  & \multicolumn{1}{c|}{85.40}   & 1.50  & 0.60     & 1.50     & 2.40   & 1.50    &                      &                      &                      &                      \\
\multicolumn{1}{c|}{CycleGAN\cite{zhu2017unpaired}}                & 82.47 & 76.94    & 78.28    & 78.89  & \multicolumn{1}{c|}{79.15}   & 1.68  & 1.35     & {1.90}     & 1.94   & 1.71    & 87.33 & 88.39    & 74.28    & 80.91  & \multicolumn{1}{c|}{82.72}   & 2.05  & 2.59     & 1.84     & 2.87   & 2.34    &                      &                      &                      &                      \\
\multicolumn{1}{c|}{DSAN\cite{DSAN}}                    & -     & -        & -        & -      & \multicolumn{1}{c|}{-}       & -     & -        & -        & -      & -       & 89.30 & 90.16    &\textbf{90.09}    & 89.84  & \multicolumn{1}{c|}{89.84}   & -     & -        & -        & -      & -       &                      &                      &                      &                      \\
\cline{1-22}
\multicolumn{1}{c|}{\textbf{Ours}}   &         \textbf{93.49}        & \textbf{85.32} &   \textbf{88.79}  &  \textbf{93.92} & \multicolumn{1}{c|}{\textbf{90.38}}  
   & 1.33  & 3.07     & 1.53     & 1.63   & 1.89    & \textbf{90.67} & \textbf{91.13}    & 89.35    & \textbf{93.02}  & \multicolumn{1}{c|}{\textbf{91.04}}   & \textbf{1.10}  & \textbf{0.41}     & \textbf{0.49}     & \textbf{0.39}   & \textbf{0.60}    &                      &                      &                      &                      \\ \cline{1-21}
\end{tabular}

\label{abdominaltable}
\end{table*}

\begin{table}[]
\centering
\caption{Ablation Study of Key Components in Cardiac MRI - CT.}
\setlength\tabcolsep{2pt}
\begin{tabular}{lcccccccc}\hline
StarGAN-v2         &   -    &    -   &    -   &  -     &    -        &  \checkmark     &   -    &     -  \\
CycleGAN           &     -  &    -   &     -  &   -    & -        &   -    &  \checkmark     &    -   \\
Style-Transfer &    -   &  \checkmark     &  \checkmark     &  \checkmark     &   -      &   -    &   -    & \checkmark     \\
$D_1$          &   -    &    -   &  \checkmark     &    -   &  \checkmark     &  \checkmark         &  \checkmark     &  \checkmark     \\
$D_2$          &    -   &   -    &    -        &  \checkmark     &  \checkmark     &  \checkmark     &  \checkmark     &  \checkmark     \\\hline

Dice(\%)           & 17.12 & 58.35 & 78.51 & 60.52 & 68.76 & 70.70 & 81.17 & \textbf{82.42} \\
          \hline
\end{tabular}

\label{ablation}
\end{table}

\begin{figure*}
  \centering
  \includegraphics[scale=0.65]{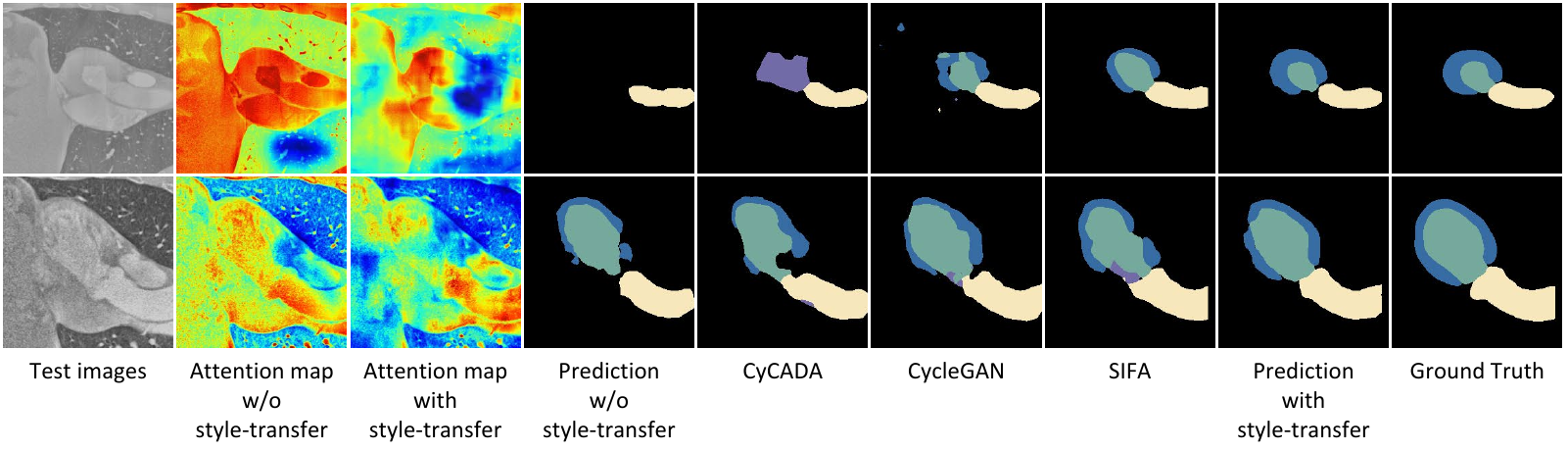}
  \caption{Visualization of the relation between local feature mask and transferability in different methods.}
  \label{fig:attentionmap}
\end{figure*}
\subsection{Analysis}
\subsubsection{Effectiveness of Key Components} To verify the influence of each module in our network, we conduct ablation tests on cardiac MRI$\rightarrow$CT segmentation task, as reported in Table~\ref{ablation}. With the style-transfer pre-processing, the Dice score raises 41.23$\%$ compared with no adaptation. The dual discriminators of local alignment module bring an improvement from 17.12$\%$ to 68.76$\%$ in Dice. Moreover, $D_1$ improves $20.16\%$ and $D_2$ improves $2.17\%$ when individually adopted with the style-transfer module. If two modules are combined, the final outcome can achieve an astonishing result of 82.42$\%$.

Besides demonstrating the effectiveness of each module and their mixtures by numerical values, we provide the visualization of the local feature mask, serving as the indicator for local feature domain gap in Fig.~\ref{fig:attentionmap}. We can infer that 1) the style-transfer step can significantly reduce the domain gap globally, especially in the background. By comparing the feature mask of the second and third columns, in the absence of global alignment, the discriminator tends to output high uncertainty in the background and considers it as part of discriminative region, thus the vital segmentation target obtains relative low scores. In this situation, the local alignment module is meaningless as aligning the background area will not contribute to final prediction. However, when the style-transfer step is adopted, the third column indicates the downstream modules can better capture discriminative features in segmentation target. 2) the discriminator $D_1$ could provide accurate feedback on the region with larger domain gap, and by adaptively re-weighting the feature map, prediction performance is improved accordingly. 
When comparing the feature mask with the SOTA results,  we observe a peak in the feature mask regions that corresponds to the areas where SOTAs do not predict accurately.
This shows our module is able to locate the real discriminative regions precisely. Then, if we add our prediction in comparison, our method performs apparently better in these areas, indicating the effectiveness of our local alignment module.


\begin{figure}
  \centering
  \includegraphics[scale=0.3]{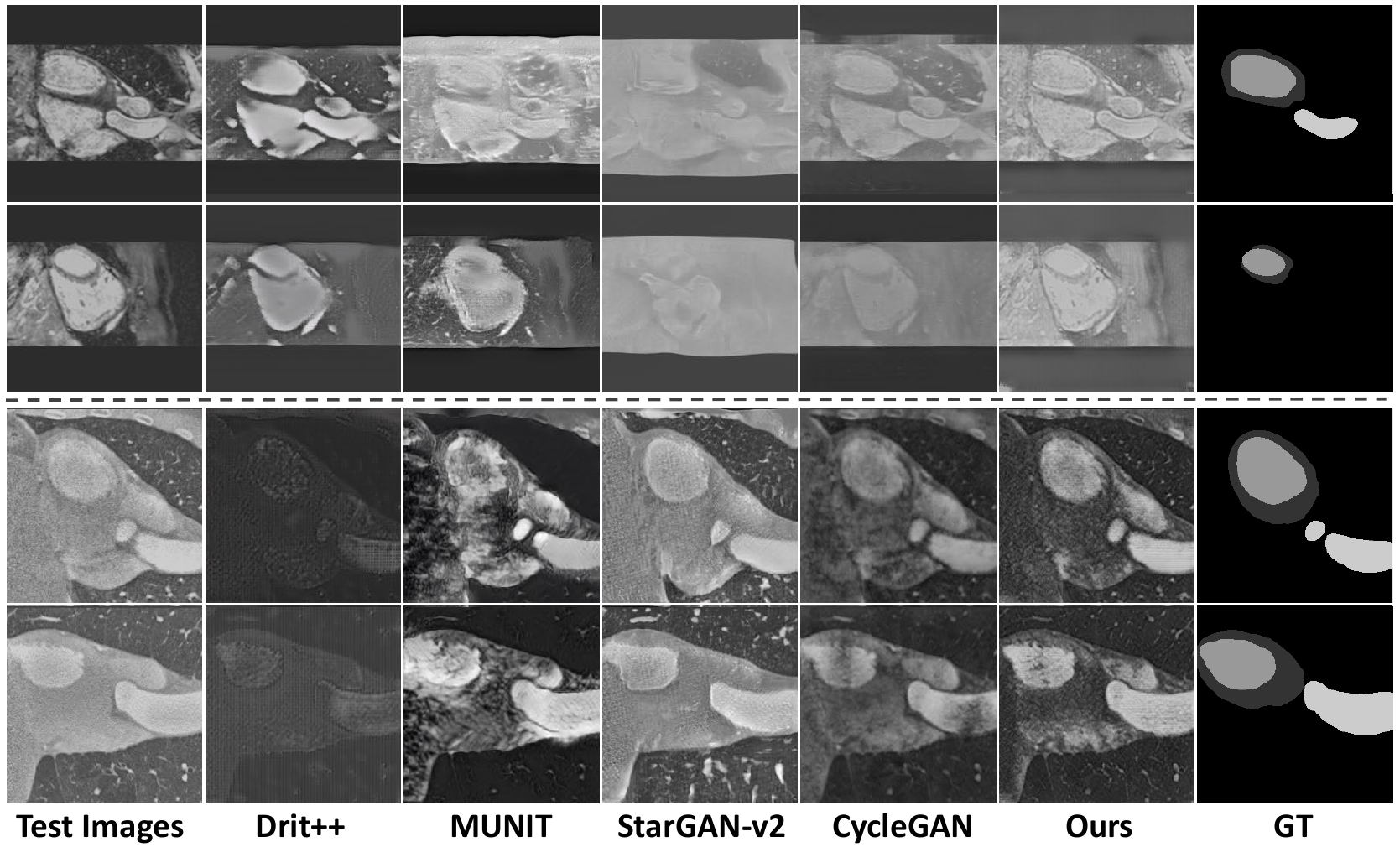}
  \caption{Comparison of style-transfer. The first two rows are MRI$\rightarrow$ CT examples. The last two rows are CT $\rightarrow$ MR examples}
  \label{fig:sss}
\end{figure}

\subsubsection{Choice of Style Transfer} We provide visualization examples in Fig.~\ref{fig:sss} to compare our style-transfer methods with other SOTA style-transfer methods~\cite{zhu2017unpaired,MUNIT,lee2020drit++,STARGAN2}. The aim of style-transfer pre-processing is to reduce the domain gap from a stylistic perspective globally, weakening the influence of different backgrounds on the subsequent local alignment. The downstream discriminator would better focus on the segmentation target. Therefore, we design the network according to the following criteria: 1) the synthesized image should be indistinguishable in style from the target domain, and 2) the semantic content of domain should be well preserved. Although it is hard to define the explicit content/style in different domains due to the deformation of structure appearance in our datasets, we assume source domain and target domain share the same content but different styles. The domain-invariant feature is taken as content that reveals the target's spatial structure, and the rendering of the structure is regarded as the style.

Visualization of different style transfer methods shows that 1) Compared with CycleGAN~\cite{zhu2017unpaired}, our style-transfer method utilizing a feature-disentangling strategy can better preserve the semantic information, revealing the effectiveness of the disentanglement; 2) Methods working well on natural image data fail to obtain satisfactory results. We believe that the methods with too strong style-transfer ability are not applicable to medical image datasets, partially due to the difficulty distinguishing the style and content in medical images. For MUNIT~\cite{MUNIT}, Drit++~\cite{lee2020drit++} and StarGAN-v2~\cite{STARGAN2}, structure deformation is quite likely as they use a sample-level training strategy. The large intra-domain variance adds the risk of negative transfer in this setting. Instead, utilizing an average style for an entire domain seems relatively safe in our scenario. 

Based on the qualitative results obtained, we further examine quantitatively two style transfers, e.g. StarGAN-v2~\cite{STARGAN2} and CycleGAN~\cite{zhu2017unpaired} on how they affect subsequent segmentation. 
The results are reported in Table~\ref{ablation}. As observed, our method still achieves the best performance.\\


\subsubsection{Limitation} 
Strictly from the perspective of feature disentangling, the shape variance of the same structure in cross-modality datasets should be taken as part of domain-specific features.  Our strategy is to utilize the average style of a domain to attain relatively stable results. This method is valid in our experiment, as the same substructure even shows diverse shapes in the same domain in our tasks. We suspect the network could notice this variation when picking the average style. However, supposing that a single class has a fixed shape in one domain and shows different morphology in another, this method is no longer appropriate. This is because it is likely to regard its shape change as style difference between domains. It consequently produces deformed images and impacts the subsequent segmentation negatively.

\section{Conclusion}
In this paper, we propose a novel strategy to better solve the domain adaptation problems in medical image datasets. By jointly aligning the domains globally and locally, the model can not only reduce the global domain shift, but also alleviate the inter-gap problem for local features. The local alignment could calibrate the local transferability, especially in crucial areas in segmentation target, thus promoting the overall performance. {We validate the proposed framework with two challenging tasks, cardiac and abdominal segmentation, and the experimental results achieve comparable even superior performance to the state-of-the-art domain adaptation methods.}

\bibliographystyle{IEEEtran.bst}
\bibliography{eccv2022submission}
\end{document}